\documentclass[runningheads]{llncs}
\usepackage{graphicx}
\usepackage[utf8]{inputenc} 
\usepackage[T1]{fontenc}    
\usepackage{hyperref}       
\usepackage{url}            
\usepackage{booktabs}       
\usepackage{amsmath}
\usepackage{amsfonts}       
\usepackage{nicefrac}       
\usepackage{microtype}      
\usepackage[table]{xcolor}  
\usepackage{acronym}
\usepackage{cleveref}
\usepackage{multirow}

\usepackage{rotating}       

\usepackage[htt]{hyphenat} 

\newcommand{\figemph}[1]{\texttt{#1}} 

\newacro{AutoML}{automated machine learning}
\newacro{NDCG}{normalized discounted cumulative gain}
\newacro{MCC}{Matthews correlation coefficient}
\newacro{FPR}{false positive rate}
\newacro{tau}[$\tau$]{Kendalls rank correlation coefficient}
\newacro{CASH}{combined algorithm selection and hyperparameter optimization}

\begin{document}
\title{Meta-Learning for Automated Selection of Anomaly Detectors for Semi-Supervised Datasets}
\titlerunning{Meta-Learning for Automated Selection of Anomaly Detectors}
%
\author{David Schubert\inst{1} \and
Pritha Gupta\inst{2} \and
Marcel Wever\inst{3}
}
\authorrunning{D. Schubert et al.}
%
\institute{Software Engineering and IT Security, Fraunhofer IEM, Paderborn, Germany \and
Department of Computer Science, Paderborn University, Paderborn, Germany \and
MCML, LMU Munich, Munich, Germany
}
\maketitle              
\begin{abstract}
	In anomaly detection, a prominent task is to induce a model to identify anomalies learned solely based on normal data. Generally, one is interested in finding an anomaly detector that correctly identifies anomalies, i.e., data points that do not belong to the normal class, without raising too many false alarms.
    Which anomaly detector is best suited depends on the dataset at hand and thus needs to be tailored.
    The quality of an anomaly detector may be assessed via confusion-based metrics such as the \ac{MCC}. However, since during training only normal data is available in a semi-supervised setting, such metrics are not accessible.
	To facilitate automated machine learning for anomaly detectors, we propose to employ meta-learning to predict \ac{MCC} scores based on metrics that can be computed with normal data only. First promising results can be obtained considering the hypervolume and the false positive rate as meta-features.
\end{abstract}

\section{Introduction}\label{sec:introduction}
\acresetall 

\ac{AutoML} \cite{hutter2019automated} refers to the vision of automating the data science process emerging from the unmatched demand for expertise in data science and machine learning in particular. A substantial part of AutoML literature is concerned with the selection, configuration, and composition of machine learning algorithms tailored for a certain task, consisting of a dataset and a performance measure \cite{zoller2021benchmark}. After first promising results for standard (binary or multi-class) classification and regression tasks on tabular data could be achieved \cite{thornton2013auto,feurer2015efficient,mohr2018ml}, AutoML quickly spread to further data types, learning problems, and tasks, such as image classification \cite{ren2021comprehensive}, multi-label classification \cite{wever2021automl}, remaining useful lifetime estimation in predictive maintenance \cite{tornede2021coevolution}, online learning \cite{celik2022online}, natural language processing \cite{bisong2019google}, and multi-modal data \cite{mueller2020faster}. While the aforementioned extensions of AutoML require supervision in terms of labels being provided for fitting the models, more recently, AutoML methods for dealing with unsupervised or semi-supervised machine learning tasks are proposed as well.

One such problem class deals with the detection of outliers, sometimes also referred to as anomalies. While performance complementarity can also be observed for anomaly detectors, i.e., for different datasets, different anomaly detection methods perform best, this setting imposes severe challenges for adopting AutoML methods to determine the most suitable anomaly detection method. First, if the provided dataset contains both classes at all, i.e., normal and anomalous, their frequency is typically unbalanced. More specifically, it is usually assumed that anomalies occur only with very low frequency. Second, the dataset at hand may comprise no anomalies at all, and the anomaly detection method is meant to detect any new data points that deviate from the training data and thus represent anomalies \cite{chandola2009anomaly,ahmed2016survey}, e.g., in intrusion detection.
In the following, we refer to this setting as being semi-supervised.
For commonly available sampling-based AutoML approaches, the latter data situation is difficult to handle since they rely on probing sampled algorithms and hyperparameter values for the given data. For example, a solution candidate is evaluated via cross-validation and the average
performance for a performance measure of interest is observed, e.g., error rate for classification or mean squared error for regression. Obviously, in the domain of anomaly detection, we are interested in finding a model that can detect anomalies reliably but also classifies normal data points as such. Performance measures such as \ac{MCC} or the area under the ROC curve are considered to account for the imbalance between normal and anomalous data points. However, these performance measures require the presence of both types of data points normal \emph{and} anomalous data points such that they cannot be computed in the semi-supervised setting mentioned above.

In this paper we investigate the use of meta-learning to overcome the issue of evaluating algorithms and their hyperparameter values for a performance measure of interest, which may require data points of both classes.
To this end, we assess the predictive power of two metrics that can be computed with normal data points only: hypervolume of anomaly detectors and the \ac{FPR}.
In our experimental section, we find promising results when utilizing these metrics in terms of meta-features for landmarking and feature descriptions of anomaly detectors.

\section{CASH for Semi-Supervised Anomaly Detection}\label{sec:problem-definition}
Let $\mathcal{X}$ and $\mathcal{Y} = \{0,1\}$ be an instance space and a target space respectively. Furthermore, we assume instances $x \in \mathcal{X}$ to be (non-deterministically) associated with a label $y \in \mathcal{Y}$. In the setting of anomaly detection, we associate instances $x$ with $y=1$ in case it is an anomaly and $y=0$ if it is normal. Moreover, anomalies are assumed to occur only seldomly, i.e., $y=1$ is a minority class, and during training only normal data points are provided. This setting is also occasionally referred to as semi-supervised anomaly detection~\cite{chandola2009anomaly}.
For convenience, let $\{0\} =: \mathcal{Y}_0 \subset \mathcal{Y}$. We seek to learn a hypothesis $h: \mathcal{X} \rightarrow \mathcal{Y}$ from training data
given in the form of $\mathcal{D} = \{ (x_i, y_i) \in \mathcal{X} \times \mathcal{Y}_0 \mid 1 \leq i \leq n \}$ that is able to discriminate well between normal and anomalous data points.

Let be $\mathcal{A} := \{ A^{(1)}, A^{(2)}, \ldots A^{(m)} \}$ a set of anomaly detectors with corresponding hyperparameter spaces $\Lambda^{(1)}, \Lambda^{(2)}, \ldots \Lambda^{(m)}$. Additionally, a dataset of training instances $\mathcal{D} := \{(\mathcal{X}, \mathcal{Y}_0)\}_{i=1}^{n}$, and a performance measure $m: \mathcal{Y} \times \mathcal{Y} \rightarrow \mathbb{R}$, we aim to find the most suitable anomaly detector $A^\ast$ together with its hyperparameter setting $\lambda^\ast$ with respect to $m$:
\begin{equation}\label{eq:automl}
A^\ast_{\lambda^\ast} \in \underset{A^{(i)} \in \mathcal{A}, \lambda \in \Lambda^{(i)}}{\arg \max} \mathbb{E}_{(x,y) \sim P}\,\, \left[ m(A^{(i)}_\lambda(\mathcal{D}, x), y) \right] \,\,\, ,
\end{equation}
where $A_\lambda^{(i)}$ is trained on training data $\mathcal{D}$ and makes a prediction on $x$ which is then compared to the ground truth $y$ and without loss of generality $m$ is to be maximized. This problem is better known as the combined algorithm selection and hyperparameter optimization (CASH) problem as initially formalized in \cite{thornton2013auto}.

To find such an $A^\ast_{\lambda^\ast}$, in sampling-based AutoML systems, one would split the dataset $\mathcal{D}$ into subsets of training and validation data $\mathcal{D}_{train}$ and $\mathcal{D}_{val}$ respectively and use the performance on the validation data as an estimate of the true generalization performance:
\begin{equation}\label{eq:practical-automl}
	\widehat{A}^\ast_{\widehat{\lambda}^\ast} \in \underset{A^{(i)} \in \mathcal{A}, \lambda \in \Lambda^{(i)}}{\arg \max} \mathbb{E}_{\mathcal{D}_{train}, \mathcal{D}_{val} = (x_j, y_j)_{j=1}^{s}}\,\, \left[ \frac{1}{s} \sum_{j=1}^{s} m(A^{(i)}_\lambda(\mathcal{D}_{train}, x_j), y_j) \right] \,\,\, .
\end{equation}

\section{Related Work}\label{sec:related-work}

To the best of our knowledge, there are no other publications that focus on \ac{CASH} in a semi-supervised setting.
Therefore, we refer to unsupervised approaches and semi-supervised hyperparameter optimization approaches in this section.

In~\cite{zhao2021automatic}, Zhao et al. present MetaOD, an approach to unsupervised outlier model selection.
It is based on the construction of meta-features for a corpus of training datasets with outlier labels and the performance of over $300$ models for each dataset.
Similarly to our evaluation, Zhao et al. use the meta-features and performance values to train a performance predictor, which is used to choose high performing models for new datasets.
However, we focus on the semi-supervised setting of outlier detection rendering landmark features meant to capture information regarding potential outliers not applicable.

Putina et al. introduce AutoAD~\cite{putina2022autoad}, a framework with the same purpose as MetaOD but without utilizing meta-learning.
They measure the performance of an anomaly detector via metrics applied to the data before and after removing the top-ranked anomalies predicted by the detector in question.
Again, this approach is not applicable to our semi-supervised setting as the employed metrics imply the existence of anomalies in the tails of the data distribution.
We do not assume any anomalies being present in the datasets for which an anomaly detector is to be optimized.
Other approaches like PyODDS~\cite{li2020pyodds}, TODS~\cite{lai2021tods}, AutoOD~\cite{li2022automated}, and LSCP~\cite{yue2019lscp} utilize supervised performance metrics rendering them not applicable to our semi-supervised setting too.

Tax et al. have a body of works evolving from the idea of the \emph{hypervolume} of one-class classifiers as being the hypervolume of the subspace that a one-class classifier predicts to belong to the target class~\cite{TM04,TD01optimization,TD01generation}.
In~\cite{TD01optimization}, they aim at optimizing the hyperparameters of the support vector data description.
For this purpose, they estimate the hypervolume of the support vector data description and use a linear combination of this estimation and the \ac{FPR} (error on the target class) as performance metric for optimization.
The linear combination encompasses a trade-off parameter that has to be set manually.

Our approach is motivated by this idea to a large extend.
However, we focus on the more general \ac{CASH} problem.
To this end, we evaluate the utilization of the hypervolume and the \ac{FPR} as features description of anomaly detectors and for landmarking purposes.
Furthermore, we train a meta-model for the prediction of the quality of anomaly detectors on unseen datasets.

\section{Meta-Learning for Selecting Anomaly Detectors}\label{sec:approach}

According to the assumption that $\mathcal{D}$ only contains normal data, i.e., data points of the form $(x_j, y_j)$ with $y_j \in \mathcal{Y}_0$, performance measures that quantify how accurately an anomaly detector may identify anomalies, e.g., via \ac{MCC}, precision, recall, or AUC, cannot be assessed for evaluation. Hence, such measures can neither be employed by AutoML systems to search for the most appropriate anomaly detector for a given data set.

To overcome this issue, we aim to substitute the performance measure $m$ in Eq.~\ref{eq:practical-automl} by a surrogate model $\widehat{m}$, which provided a feature description $f_A \in \mathbb{R}^k$ of algorithm $A^{(i)}$ together with its hyperparameter setting $\lambda$ and a feature description $f_\mathcal{D} \in \mathbb{R}^l$ of the dataset $\mathcal{D}$ in question predicts the measure of interest. To this end, the surrogate is build on data sets for which anomalies are actually known and hence a performance value can be computed. Then, we can substitute $m$ by $\widehat{m}$ in Eq.~\ref{eq:practical-automl} to obtain
$$
\widehat{A}^\ast_{\widehat{\lambda}^\ast} \in \underset{A^{(i)} \in \mathcal{A}, \lambda \in \Lambda^{(i)}}{\arg \max}\,\, \widehat{m}(f_\mathcal{A}(A^{(i)}, \lambda), f_\mathcal{D}(\mathcal{D})) \,\,\, .
$$

In the following we consider mainly two types of features as to describe algorithms as well as datasets (in terms of landmarking features): hypervolume and \ac{FPR}.
\begin{description}
\item[Hypervolume] The hypervolume can be considered as a means to describe how tightly an anomaly detector fits the normal data points. With a smaller hypervolume, chances are low that anomalies are missed, whereas a larger hypervolume may lead to anomalies not being identified as such.
\item[False Positive Rate] The \ac{FPR} assesses how many data points in the training data are falsely classified as anomalies, i.e., the anomaly detector would raise a false alarm in this case.
\end{description}
It is easy to see that minimizing both the hypervolume and the \ac{FPR} is conflicting with each other. Therefore, we seek to find a method yielding a suitable tradeoff between the two measures.

\section{Experimental Evaluation}
\label{sec:Evaluation}

In the following we assess the predictive power of the hypervolume and the \ac{FPR} for the task of selecting a suitable anomaly detector for an unseen dataset.
First, we consider a linear combination of the features to order anomaly detectors and evaluate the general usefulness of these metrics in terms of a feature description of the detectors.
Second, we train a meta-model to predict the \ac{MCC} score of anomaly detectors. This model gets as an input landmarking features resulting from the hypervolume and \ac{FPR} of a fixed portfolio of anomaly detectors and the hypervolume and \ac{FPR} of the anomaly detector in question.

\begin{figure}[t]
	\centering
	\includegraphics[width=.9\textwidth]{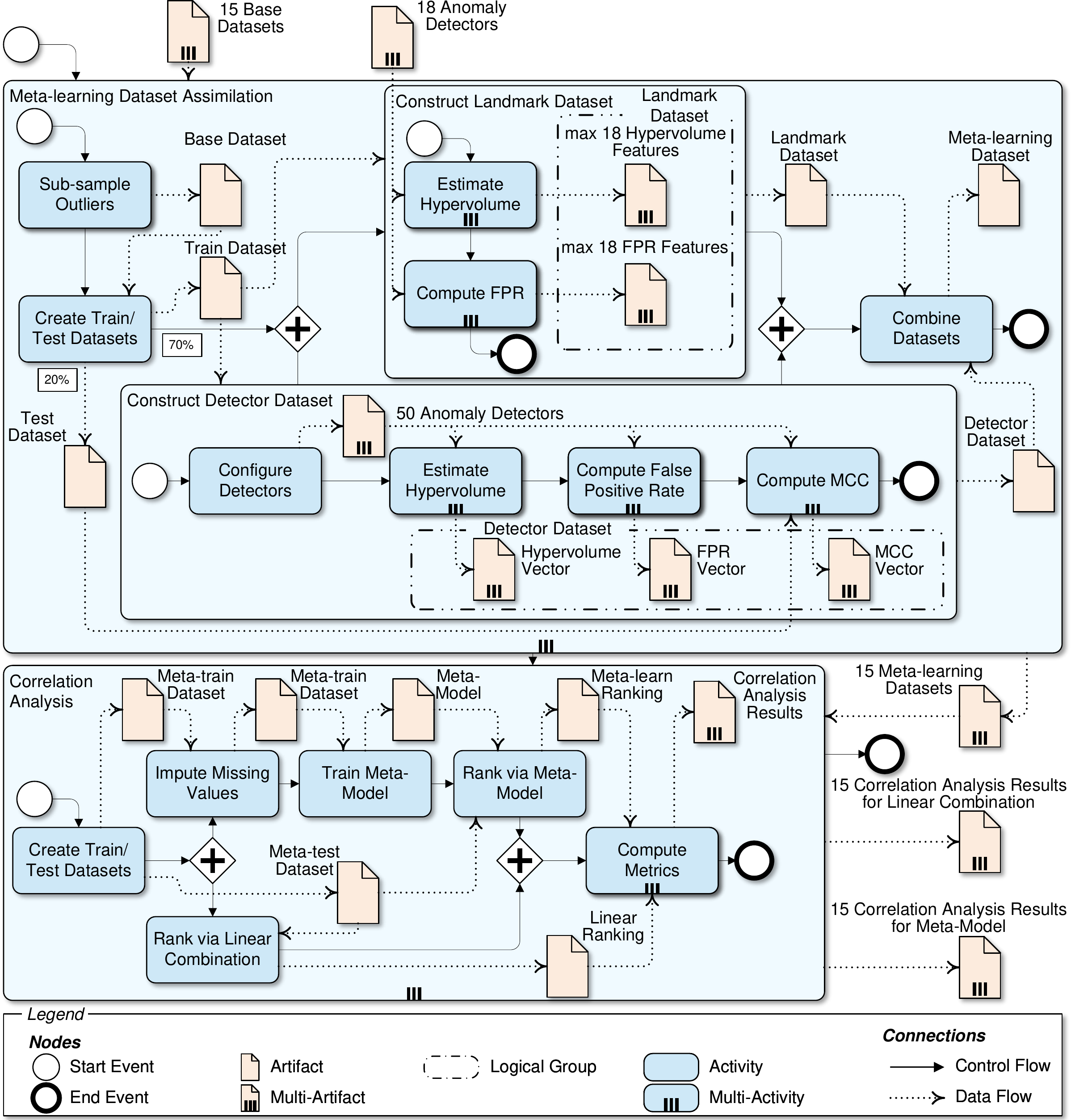}
	\caption{BPMN Diagram~\cite{BPMN} of the Experiments}
	\label{fig:ExperimentSetup}
\end{figure}

Figure~\ref{fig:ExperimentSetup} shows the process that we follow for our experiments.
We continuously refer to this figure in the course of this section.
All experiments of the evaluation were executed on virtual machines with Ubuntu 20.04.4, 16 cores (Intel Xeon E5-2695 v3), and 128 GB RAM.
Sections~\ref{sec:dataset:assimilation} and \ref{sec:correlation_analysis} elaborate on the assimilation of the meta-learning dataset and a correlation analysis respectively.
Thereafter, Section~\ref{sec:results} discusses the results of the evaluation.

\subsection{Meta-Learning Dataset Assimilation}
\label{sec:dataset:assimilation}
We base our meta-learning datasets on a number of datasets for outlier detection.
The source of these datasets is the collection of Tax et al.~\cite{datasets}.
The estimation of the hypervolume of anomaly detectors, which we explain in the context of the corresponding activity, scales poorly with an increasing number of dimensions.
Thus, we only select those datasets that have 8 features or less, resulting in 15 \figemph{Base Datasets}, which can be inferred from Table~\ref{tab:results}.
We execute the \figemph{Meta-learning Dataset Assimilation} activity iteratively for all \figemph{15 Base Datasets}.

One of our basic assumptions is that outliers are the minority class (cf. Section~\ref{sec:problem-definition}).
Not all of the datasets reflect this assumption.
Thus, we \figemph{Sub-sample Outliers} to ensure that anomalies constitute between 5\% and 10\% of the total dataset.
We do this only for those datasets, where outliers originally amount to more than 10\% of the dataset.
To \figemph{Create the Train and Test Datasets}, we shuffle the \figemph{Base Dataset} and perform a stratified split into \figemph{Train} (70\%) and \figemph{Test Datasets} (20\%).
The remaining 10\% are kept for future experiments that are not in the focus of this publication.
Furthermore, we remove all outliers from the \figemph{Train dataset} to match our semi-supervised setting.
Lastly, we scale the \figemph{Train} and \figemph{Test Datasets}, subtracting the mean and scale the features independently to the interquartile range with respect to the \figemph{Train Dataset}.

Figure~\ref{fig:ExperimentSetup} shows, that the control flow forks at this point into the \figemph{Construct Landmark Dataset} and \figemph{Construct Detector Dataset} activities.
These activities extract our meta-features, which are meant to characterize datasets and anomaly detectors, respectively.

The landmark features are the hypervolume and the \ac{FPR} of the \figemph{Anomaly Detectors} provided by PyOD\footnote{\url{https://pyod.readthedocs.io/} in version 0.9.7}~\cite{zhao2019pyod} in their default configuration.
Here, we exclude those algorithms that do not finish after one week of runtime for each dataset resulting in at most \figemph{18 Anomaly Detectors} per \figemph{Base Dataset}.

The estimation of the hypervolume that a model spans with respect to a certain dataset follows the approach of Tax et al.~\cite{TD01generation}.
Correspondingly, we fit the smallest hyper-sphere around the train data and generate 35 million uniformly distributed data instances within this sphere.
We train the \figemph{Anomaly Detectors} on the \figemph{Train Dataset}.
The fraction of instances classified as belonging to the target class is an estimate of the hypervolume of the corresponding anomaly detector.
Please note, that we inferred the number of generated data instances from initial experiments.
Generally, the number of necessary data instances grows exponentially with the number of dimensions of the \figemph{Train Dataset}.
This is the reason for the aforementioned bad scaling of the hypervolume estimation.
To \figemph{Compute the \ac{FPR}}, we execute a monte-carlo cross validation with a 30\% test-size and ten repetitions on the \figemph{Train Dataset}.

The parallel activity \figemph{Construct Detector Dataset} extracts the hypervolume, the FPR, and the \ac{MCC} of 50 randomly configured \figemph{Anomaly Detectors} of PyOD per \figemph{Train Dataset}.
The estimation of the hypervolume of these \figemph{50 Anomaly Detectors} and the FPR follows the computations in the context of the \figemph{Landmark Dataset} explained above.
We use the \ac{MCC} of these anomaly detectors as a performance measure, which we compute on the corresponding \figemph{Test Dataset}.
If the feature extraction via a particular anomaly detector does not terminate within 5 hours (while executing 4 anomaly detectors in parallel), we replace it with a newly configured anomaly detector.

The last activity is \figemph{Combine Datasets}, which uses the \figemph{Landmark Dataset} and the \figemph{Detector Dataset} to create the \figemph{Meta-learning Dataset}.
We combine the features of the former datasets such that each instance of the resulting \figemph{Meta-learning Dataset} encompasses all landmarking features of the \figemph{Train Dataset} and the detector features of one particular anomaly detector.

\subsection{Correlation Analysis}
\label{sec:correlation_analysis}

We execute the \figemph{Correlation Analysis} iteratively for all \figemph{15 Meta-learning Datasets}.
The \figemph{Meta-train} and \figemph{Meta-test Datasets} are created in a leave-one-out fashion.
Meaning, in each iteration of the \figemph{Correlation Analysis} one \figemph{Meta-learning Dataset} is used as \figemph{Meta-test Dataset} and the remainder is merged and used as \figemph{Meta-train Dataset}.
Since not all of the \figemph{18 Anomaly Detectors} terminate for all datasets in the context of the \figemph{Compute Landmark Dataset} activity, we remove all landmark features that have no values in the \figemph{Meta-test Dataset} and scale the remaining features individually to be in $[0,1]$.

After the creation of these datasets, the control flow forks again.
The lower control flow ranks the \figemph{Anomaly Detectors} covered by the \figemph{Meta-test Dataset} according to the simple linear combination of their detector features as a score:
\begin{equation}\label{eq:linear_combination}
lc(A_{\lambda}, \mathcal{D}) = 1-\frac{hypervolume(A_{\lambda}, \mathcal{D})+FPR(A_{\lambda}, \mathcal{D})}{2} \,\,\, .
\end{equation}

The upper control flow ranks the \figemph{Anomaly Detectors} via a meta-model that predicts the \ac{MCC} of the \figemph{Anomaly Detectors} covered by the \figemph{Meta-test Dataset} based on the landmarking and detector hypervolume and \ac{FPR} features.
We opt for a mean strategy to \figemph{Impute Missing Values}.
Furthermore, we use the \figemph{Meta-train Dataset} to train a random forest regressor (\figemph{Meta-Model}) in default parameterization\footnote{\url{https://scikit-learn.org/1.1/modules/generated/sklearn.ensemble.RandomForestRegressor.html}}.
Thereafter, we rank the \figemph{Anomaly Detectors} of the \figemph{Meta-test Dataset} using the predicted \ac{MCC} of the \figemph{Meta-Model} as a score.

We execute the \figemph{Compute Metrics} activity for the \figemph{Linear Ranking} and the \figemph{Meta-learn Ranking} separately.
At the beginning of the activity, we scale the \ac{MCC} to the $[0,1]$ interval to avoid coping with negative scores.
As metrics, we use the regret@k, \ac{tau}~\cite{Knight66}, and the \ac{NDCG}~\cite{cumulated2002jarvelin}, which are metrics that refer to rankings.
We base our evaluation on ranking metrics because a correct ranking of anomaly detectors is more important to guide an \ac{AutoML} system than precise prediction of an anomaly detectors performance, e.g., in terms of the mean squared error.
While a high precision in the predicted quality is certainly desirable, it is not necessary for choosing the most promising out of a set of anomaly detectors.
Here, a precise ranking is sufficient.

The regret@k compares the performance of the best model within a top-k ranking with the actual best model known for a dataset and reports the absolute difference.
In our case it refers to the scaled \ac{MCC} and gives an intuition about the performance of the top-ranked anomaly detectors. 

\ac{tau} measures the correspondence between two rankings, which are the ranking given by the method in question and the optimal ranking given by the true \ac{MCC} values.
\ac{tau} ranges in the interval $[-1,1]$ where negative values indicate a negative correlation of the rankings and positive values a positive correlation.
We utilize the b-version of \ac{tau}, which accounts for ties.

Similarly, the \ac{NDCG} is a measure to compare a predicted ranking to an optimal one.
In comparison to \ac{tau}, the actual scores (in our case the scaled \ac{MCC} values) influence the \ac{NDCG} and not only the ranking inferred from those scores.
The \ac{NDCG} ranges in the $[0,1]$ interval where values close to 1 denote a high quality of the predicted ranking. 

\subsection{Results}
\label{sec:results}
In this section, we discuss the performance of the \figemph{Linear Ranking} and \figemph{Meta-learn Ranking} approaches discussed in Section~\ref{sec:correlation_analysis}. 
Due to the lack of directly comparable approaches (cf. Section~\ref{sec:related-work}), we introduce three different baselines.
Table~\ref{tab:results} reports on the performance of the approaches with respect to the metrics introduced in Section~\ref{sec:correlation_analysis}.
Additionally, it reports the maximum, mean, and minimum of the \ac{MCC} of the anomaly detectors considered for prediction scaled to $[0,1]$ to give an impression of the corresponding distribution.
Please note that the \ac{NDCG} is generally not particularly insightful and reported for completeness.

\begin{sidewaystable}
  \centering
  \caption{Correlation analysis results. Small arrows indicate whether a metric is to be maximized or minimized where the best performance per dataset and metric is bolded. For datasets marked with a $^\dagger$, anomaly detectors considered for prediction have a mean scaled \ac{MCC} score of at least 0.6, meaning they actually are capable of identifying anomalies reasonably. We use the following abbreviations for the ranking approaches: R = random, FPR = false positive rate, HV = hypervolume, L = linear combination, M = meta-model, and Mc = cherry picked variant of the meta-model.}
  \label{tab:results}
  \rowcolors{4}{white}{gray!25}
    \begin{tabular}{l|ccc||ccc|ccc||ccc|ccc||ccc|ccc||ccc|ccc}
       \toprule
        Meta-test
        & \multicolumn{3}{c||}{scaled \ac{MCC}} & \multicolumn{6}{c||}{regret@1 $\downarrow$}     &  \multicolumn{6}{c||}{regret@5 $\downarrow$}& \multicolumn{6}{c||}{\ac{NDCG} $\uparrow$} & \multicolumn{6}{c}{\ac{tau} $\uparrow$} \\
        \cmidrule{2-28}
        Dataset & \scriptsize{Max} & \scriptsize{Mean} & \scriptsize{Min} & \scriptsize{R} & \scriptsize{FPR} & \scriptsize{HV} & \scriptsize{L} & \scriptsize{M} & \scriptsize{Mc} & \scriptsize{R} & \scriptsize{FPR} & \scriptsize{HV} & \scriptsize{L} & \scriptsize{M} & \scriptsize{Mc} & \scriptsize{R} & \scriptsize{FPR} & \scriptsize{HV} & \scriptsize{L} & \scriptsize{M} & \scriptsize{Mc} & \scriptsize{R} & \scriptsize{FPR} & \scriptsize{HV} & \scriptsize{L} & \scriptsize{M} & \scriptsize{Mc}  \\
        \midrule
\scriptsize{Balance Middle}   &  .59 &  .47 &  .40 &                    .12 &            .11 &      \textbf{.07} &               .12 &              .12 &                 &                    .05 &            .10 &               .05 &      \textbf{.03} &              .04 &                 &                   .96 &  \textbf{.97} &     \textbf{.97} &              .96 &             .96 &                &             -.01 &  \textbf{.15} &          -.16 &          -.01 &          -.14 &             \\
\scriptsize{Biomed Healthy}   &  .84 &  .45 &  .31 &                    .35 &            .38 &      \textbf{.30} &               .42 &              .38 &                 &           \textbf{.22} &            .38 &               .30 &               .42 &              .38 &                 &          \textbf{.89} &  \textbf{.89} &     \textbf{.89} &              .86 &    \textbf{.89} &                &              .01 &  \textbf{.22} &          -.26 &          -.22 &           .06 &             \\
\scriptsize{Diabetes Absent}  &  .67 &  .56 &  .45 &                    .07 &            .18 &               .13 &               .19 &     \textbf{.04} &                 &                    .03 &            .12 &      \textbf{.02} &               .12 &              .03 &                 &                   .96 &           .93 &     \textbf{.97} &              .94 &    \textbf{.97} &                &              .01 &          -.38 &  \textbf{.26} &          -.23 &          -.06 &             \\
\scriptsize{Diabetes Present} &  .67 &  .51 &  .41 &                    .16 &            .19 &               .17 &               .22 &     \textbf{.14} &                 &                    .09 &            .19 &      \textbf{.07} &               .22 &              .14 &                 &                   .95 &           .94 &              .95 &              .93 &    \textbf{.96} &                &              0 &          -.11 &          -.26 &          -.36 &  \textbf{.03} &             \\
\scriptsize{Ecoli Periplasm}  &  .73 &  .57 &  .40 &                    .15 &   \textbf{.01} &               .17 &               .17 &              .20 &                 &                    .07 &   \textbf{.01} &               .15 &               .11 &              .05 &                 &                   .95 &  \textbf{.96} &              .95 &              .95 &             .94 &                &             -.01 &           .03 &           .13 &  \textbf{.16} &          -.01 &             \\
\scriptsize{Iris Versicolor}  &  .73 &  .49 &  .36 &                    .26 &            .26 &      \textbf{.14} &               .26 &              .30 &                 &                    .13 &   \textbf{0} &               .10 &               .12 &              .12 &                 &                   .93 &           .93 &     \textbf{.96} &              .94 &             .92 &                &             -.02 &          -.19 &  \textbf{.28} &           .13 &          -.06 &             \\
\scriptsize{Iris Virginica}   &  .84 &  .53 &  .35 &                    .24 &            .37 &               .21 &      \textbf{0} &              .37 &                 &                    .14 &   \textbf{0} &               .13 &      \textbf{0} &     \textbf{0} &                 &                   .91 &           .89 &     \textbf{.96} &     \textbf{.96} &             .90 &                &              .01 &          -.11 &  \textbf{.45} &           .39 &           .17 &             \\
\scriptsize{Liver 1}          &  .64 &  .49 &  .40 &           \textbf{.14} &            .16 &               .16 &               .19 &              .16 &                 &                    .08 &            .16 &               .10 &               .10 &     \textbf{.03} &                 &                   .95 &           .95 &              .94 &              .94 &    \textbf{.96} &                &             0 &          0 &          -.20 &          -.09 &  \textbf{.04} &             \\
\scriptsize{Survival G5}      &  .85 &  .54 &  .40 &                    .31 &            .39 &               .39 &      \textbf{.19} &              .34 &                 &                    .22 &   \textbf{0} &               .25 &      \textbf{0} &              .23 &                 &                   .94 &           .95 &              .93 &     \textbf{.97} &             .93 &                &             -.02 &           .30 &           .05 &  \textbf{.41} &          -.11 &             \\
\scriptsize{Survival S5}      &  .72 &  .50 &  .37 &           \textbf{.21} &            .26 &               .27 &               .27 &              .27 &                 &                    .11 &            .24 &      \textbf{.08} &               .17 &              .26 &                 &          \textbf{.94} &           .92 &     \textbf{.94} &              .92 &             .92 &                &              .03 &          -.15 &  \textbf{.18} &          0 &          -.05 &             \\
\midrule
\scriptsize{Biomed Diseased$^\dagger$}  &  1 &  .64 &  .31 &                    .15 &   \textbf{0} &               .50 &      \textbf{0} &     \textbf{0} &      \textbf{0} &                    .09 &   \textbf{0} &               .37 &      \textbf{0} &     \textbf{0} &      \textbf{0} &                   .90 &           .97 &              .89 &              .96 &    \textbf{.98} &     \textbf{.98} &              0 &           .44 &          -.02 &           .41 &  \textbf{.54} &  \textbf{.54} \\
\scriptsize{Balance Left$^\dagger$}     &  .92 &  .62 &  .44 &                    .30 &            .33 &               .31 &      \textbf{0} &              .33 &               .13 &                    .18 &            .13 &               .30 &      \textbf{0} &              .16 &               .13 &                   .94 &           .94 &              .94 &     \textbf{.98} &             .95 &              .96 &              0 &           .07 &           .24 &  \textbf{.48} &           .29 &           .39 \\
\scriptsize{Balance Right$^\dagger$}    &  .86 &  .62 &  .42 &                    .25 &            .39 &               .23 &      \textbf{0} &              .25 &               .08 &                    .11 &            .08 &               .23 &      \textbf{0} &              .23 &      \textbf{0} &                   .94 &           .92 &              .94 &              .98 &             .93 &     \textbf{.99} &              0 &           .12 &           .12 &           .53 &           .01 &  \textbf{.59} \\
\scriptsize{Iris Setosa$^\dagger$}      &  1 &  .68 &  .41 &                    .22 &   \textbf{0} &               .42 &               .27 &              .35 &               .27 &                    .16 &   \textbf{0} &               .34 &      \textbf{0} &              .27 &      \textbf{0} &                   .93 &  \textbf{.98} &              .91 &              .95 &             .92 &              .97 &              0 &  \textbf{.46} &          -.24 &           .28 &          -.18 &           .28 \\
\scriptsize{Liver 2$^\dagger$}          &  .85 &  .60 &  .35 &                    .12 &   \textbf{0} &               .32 &      \textbf{0} &              .32 &      \textbf{0} &                    .05 &   \textbf{0} &               .18 &      \textbf{0} &              .18 &      \textbf{0} &                   .93 &  \textbf{.99} &              .91 &              .96 &             .91 &              .97 &              .02 &  \textbf{.49} &          -.17 &           .21 &          -.18 &           .23 \\
\midrule[0.3pt]\bottomrule[1pt]
\scriptsize{Mean}                 &  .79 &  .55 &  .39 &                    .20 &            .20 &               .25 &               .15 &              .24 &      \textbf{.10} &                    .12 &            .09 &               .18 &               .09 &              .14 &      \textbf{.03} &                   .93 &           .94 &              .94 &              .95 &             .94 &     \textbf{.97} &              0 &           .09 &           .03 &           .14 &           .02 &  \textbf{.41} \\
\scriptsize{Median}               &  .84 &  .54 &  .40 &                    .21 &            .19 &               .23 &               .19 &              .27 &      \textbf{.08} &                    .11 &            .08 &               .15 &               .03 &              .14 &      \textbf{0} &                   .94 &           .94 &              .94 &              .95 &             .93 &     \textbf{.97} &              0 &           .07 &           .05 &           .16 &          -.01 &  \textbf{.39} \\
    \end{tabular}
\end{sidewaystable}

\paragraph{Baselines:}

The first baseline approach is based on randomization.
Here, we report the mean values of 50 randomized rankings.
The second approach utilizes the \ac{FPR} detector feature introduced in Section~\ref{sec:dataset:assimilation} and ranks the 50 anomaly detectors per dataset with the inversed \ac{FPR} ($1-FPR(A_{\lambda})$) as score.
The last baseline approach works analogously to the second one but uses the inverse of the hypervolume detector feature.

Table~\ref{tab:results} shows the conflicting nature of the \ac{FPR} and hypervolume features already mentioned in Section~\ref{sec:approach} as good \ac{tau} performances for one approach are typically accompanied with rather poor ones for the other.
Additionally, the regret and \ac{tau} indicate that the \ac{FPR} is more informative than the hypervolume.
One possible reason for this is that the \ac{FPR} directly influences the \ac{MCC} but the hypervolume only if it correlates with the false negative rate.
This is not necessarily the case, e.g., if an anomaly detector exactly covers the hypersphere, which in turn does not cover any anomalies.
Furthermore, the hypervolume is very sensitive to noisy features.
An anomaly detector can expand in the direction of such features and drastically increase its volume without affecting its \ac{MCC}.

\paragraph{Linear Ranking:} 

Ranking according to the linear combination of hypervolume and \ac{FPR} detector features, yields decent results.
The linear combination of the hypervolume and the \ac{FPR} seems to be a suitable means to compare different anomaly detectors to each other and rank them accordingly.
Particularly when comparing the combined approach with the separate baselines introduced before, we see that the combination outperforms the separate approaches.
For 5 datasets the linear rankings have the top detector in the first place and for two more datasets it is within the top-5 ranked detectors.

Furthermore, we see that the mean and median of \ac{tau} indicate a positive association in relation to optimal rankings.
However, investigating \ac{tau} for the datasets separately yields that 5 datasets show negative \ac{tau} values, which is also the reason for the relative low mean and median \ac{tau} of $.14$ and $.16$, respectively.
Cross-checking these results with the mean \ac{MCC} values of the 50 randomly configured anomaly detectors per dataset indicates a strong relation.
If we restrict the evaluation to those datasets with a mean scaled \ac{MCC} of at least $.6$, we end up with a mean regret@1 of $0.05$, regret@5 of $0$, \ac{NDCG} of $0.97$, and \ac{tau} of $0.38$ (not shown in Table~\ref{tab:results}).
Hence, the features seem to be more informative for those datasets for which we randomly find better anomaly detectors on average.

\paragraph{Meta-learn Ranking:} 

Using the hypervolume and \ac{FPR} as meta-features for both datasets and the anomaly detector is clearly worse than the linear combination of the detector features alone.
Overall, it can be regarded as being on par with the randomized approach.

However, a leave-one-out evaluation on cherry picked datasets -- analogously to what we describe in the context of the linear ranking -- leads to the meta-model slightly outperforming the cherry picked linear ranking in terms of the mean \ac{NDCG} and \ac{tau}.
Thus, we once more see a strong relation to particularities of the corresponding meta-learning datasets and the performance of the encompassed anomaly detectors, which -- obviously -- the landmarking features cannot sufficiently express.
Additionally, the hypervolume estimation is very sensitive to the data distribution of the target class.
On the one hand, for distributions well approximated by the sampled hypersphere, a good anomaly detector has a hypervolume close to 1.
On the other hand, for a distribution on a submanifold, the hypervolume of a good detector tends towards 0.
This relation might be hard to learn.
Still, we find the results to be promising, particularly considering that we do not assume any information regarding the outlier class of the unseen dataset.
Thus, they form a potential base from which interesting future work may emerge.
Especially, whether additional meta-features, other types of meta-models, or improved sampling strategies for the hypervolume estimation may help to improve the accuracy of the predictions for the \ac{MCC}.

\section{Conclusion}

In this paper, we have considered the learning problem of anomaly detection where during training only a dataset with normal data points is available.
While this impedes the use of performance measures explicitly quantifying how well an anomaly detector is able to identify anomalies, for the task of automatically selecting and configuring anomaly detectors, we proposed to employ meta-learning to predict measures of interest.
In this regard, we considered mainly two types of meta-features.
One that is based on the hypervolume covered by trained anomaly detectors and the other one considering the \acp{FPR} of anomaly detectors. While a lower value for the former seems favorable as the anomaly detector fits the training data more tightly, this is usually in conflict with minimizing the \ac{FPR} since the smaller the hypervolume, the more training data points may be classified as positive.
Used in combination, the two features have shown promising performance to be used directly for ranking anomaly detectors and as meta-features for a meta-model to predict a measure of interest that would actually require anomalies in the training data for being evaluated.
Moreover, results of the corresponding experiments, which can rather be considered a proof of concept, indicate that AutoML for anomaly detectors might be feasible using such surrogate measures for performance evaluation.

Whether AutoML systems can really work well with such surrogate models for selecting and configuring anomaly detectors, however, is still an open question and also outlines interesting future work.
Thus, we aim to extend our approach by augmenting the set of meta-features by more types of informative features describing the data or the anomaly detector, to improve the quality of the meta-model.
Other research directions are too improve the hypervolume estimation and to formalize the problem directly as ranking problem.

\bibliographystyle{unsrt}
\bibliography{literature}

\appendix

\end{document}